\documentclass[a4paper,10pt]{article}

\usepackage{graphicx}
\usepackage{amsthm}

\begin{document}

\title{Learning Hierarchical Sparse Representations using Iterative Dictionary Learning and Dimension Reduction}
\author{Mohamad Tarifi, Meera Sitharam, Jeffery Ho}
\maketitle

\begin{abstract}
This paper introduces an elemental building block which combines Dictionary Learning and Dimension Reduction (DRDL). 
We show how this foundational element can be used to iteratively construct a Hierarchical Sparse Representation (HSR) 
of a sensory stream. We compare our approach to existing models showing the generality of our simple prescription. We then perform
preliminary experiments using this framework, illustrating with the example of an object recognition task using standard datasets.
This work introduces the very first steps towards an integrated framework for designing and analyzing various computational
tasks from learning to attention to action. The ultimate goal is building a mathematically rigorous, integrated theory of intelligence.

\end{abstract}

\section*{Introduction}	

Working towards a Computational Theory of Intelligence, we develop a computational framework inspired by
ideas from Neuroscience. Specifically, we integrate notions of columnar organization, hierarchical structure,
sparse distributed representations, and sparse coding.

An integrated view of Intelligence has been proptosed by Karl Friston based on free-energy \cite{Friston2007, Fristona, Friston2008, Friston2009, Friston2010, Friston2011}. In this framework, Intelligence is viewed
as a surrogate minimization of the entropy of this sensorium. This work is intuitively inspired by this view, aiming to 
provide a computational foundation
for a theory of intelligence from the perspective of theoretical computer science, thereby connecting to ideas in mathematics.
By building foundations for a principled approach, the computational essence of problems can be isolated, formalized,
and their relationship to fundamental problems in mathematics and theoretical computer science can be illuminated
 and the full power of available mathematical techniques can be brought to bear.

A computational approach is focused on developing tractable algorithms. exploring the complexity limits of Intelligence.
Thereby improving the quality of available guarantees for evaluating performance
of models, improving comparisons among models, and moving towards provable guarantees such as sample size, time complexity,
generalization error, assumptions about prior.
 This furnishes a solid theoretical foundation
 which may be used, among other things, as a basis for building Artificial Intelligence.

\subsection{Background Literature In A Glance}   

Speculation on a cortical micro-circuit element dates back to Mountcastle's observation that a cortical column may serve as
an algorithmic building block of the neocortex \cite{Mountcastle}. Later work by Lee and Mumford \cite{Lee2003}, Hawkins and George \cite{George2009} attempted further investigation
of this process.

The bottom-up organization of cortex is generally assumed to be a hetrarchical topology of columns. This can be modeled
as a directed acyclic graph, but is usually simplified to a hierarchical tree.
Work by Poggio, Serre, et al \cite{Riesenhuber1999, Serre2007, Serre2007a, Serre2007b}, Dean  \cite{Dean, Dean1999}, discuss a hierarchical topology. Smale et al attempts to develop a theory accounting for the importance of
the hierarchical structure \cite{Smale2007, Bouvrie}.

Work on modeling early stages of sensory processing by Olshausen \cite{Olshausen1997, Olshausen1996} using sparse coding produced results that account for the observed
receptive fields in early visual processing. This is usually done by learning an overcomplete dictionary. However it remained unclear
how to extend this to higher layers. Our work can be partially viewed as a progress in this direction.   

Computational Learning Theory is the formal study of learning algorithms. PAC defines a natural setting for analyzing such algorithms.
However, with few notable exceptions (Boosting, inspiration for SVM, etc) the produced guarantees are divorced from practice.
Without tight guarantees Machine Learning is studied using experimental results on standard benchmarks, which is problematic.
We aim at closing the gap between theory and practice by providing stronger assumptions on the structures and forms considered by the theory,
through constraints inspired by biology and complex systems.

\subsection{A Variety of Hierarchical Models}

Several hierarchical models have been introduced in the literature.
H-Max is based on Simple-Complex cell hierarchy of Hubel and Wiesel.
It is basically a hierarchical succession of template matching and a max-operations,
corresponding to simple and complex cells respectively \cite{Riesenhuber1999}.
 
Hierarchical Temporal Memory (HTM) is a learning model composed of
a hierarchy of spatial coincidence detection and temporal pooling.
Coincidence detection involves finding a spatial clustering of the input,
while temporal pooling is about finding variable order Markov chains describing
temporal sequences in the data.

H-Max can be mapped into HTM in a straightforward manner. In HTM, the transformations
in which the data remains invariant is learned in the temporal pooling step. H-Max explicitly
hard codes translational transformations through the max operation. This gives H-Max better sample complexity
for the specific problems where translational invariance is present.

Bouvrie et al \cite{Bouvrie, Bouvrie2010} introduced a generalization of hierarchical architectures centered around a foundational element
involving two steps, Filtering and Pooling. Filtering is described through a reproducing Kernel $K(x,y)$, such as
the standard inner product $K(x,y) =\langle x, y\rangle$, or a Gaussian kernel $K(x,y) = e^{- \gamma \Arrowvert x -y \Arrowvert^2}$.
Pooling then remaps the result to a single value. Examples of pooling functions include max, mean, and $l_p$ norm
(such as $l_1$ or $l_{ \infty }$). H-max, Convolutional Neural Nets, and Deep Feedforward Neural Networks all belong to this
category of hierarchical architectures corresponding to different choices of the Kernel and Pooling functions.
 Our model does not fall within Bouvrie's present framework, and can be viewed as a generalization of hierarchical
models in which both HTM and Bouvrie's framework are a special case.

Friston proposed Hierarchical Dynamic Models (HDM) which are similar to the above mentioned architectures but framed
in a control theoretic framework operating in continuous time \cite{Friston2008}. A computational formalism of his approach is thus prohibitively difficult.

\subsection{Scope}	

Our approach to the circuit element is an attempt to abstract the computationally fundamental processes. We conjecture a class of
possible circuit elements for bottom-up processing of the sensory stream.  

Feedback processes, mediating action and attention, can be incorporated into this model, similar to work
by Chikkerur et al \cite{Chikkerur2009, Chikkerur2010}, and more generically to a theory by Friston \cite{Friston2010, Friston2009}. We choose to leave feedback for future work, allowing us to focus here on 
basic
aspects of this model.

\subsection{Contribution and Organization}

The following section introduces a novel formulation of an elemental building block
that could serve as a the bottom-up piece in the common cortical algorithm. This circuit element
combines Dictionary Learning and Dimension Reduction (DLDR). After formally introducing DRDL, we show how it 
can be used to iteratively construct a Hierarchical Sparse Representations (HSR) of a sensory stream.
Comparisons to relevant known models are presented.
To gain further insight, the model is applied to standard vision datasets. This immediately leads
to a classification algorithm that uses HSR for feature extraction. In the appendix, we discuss how further assumptions about the prior can naturally be expressed in our framework.

\section{DRDL Circuit Element}	

Our circuit element is a simple concatenation of a Dictionary Learning (DL) step followed by a Dimension Reduction (DR) step.
Using an overcomplete dictionary increase the dimension of the data, but since the data is now sparse, we can use Compressed
Sensing to obtain a dimension reduction.

\subsection{Dictionary Learning}	

Dictionary Learning obtains a sparse representation by learning features on which the data $x_i$ can be written in sparse linear combinations. 

\newtheorem{definition}{Definition}
\begin{definition} 
Given an input set $X=[x_0 \ldots x_m]$, $x_i \in R^d$, Dictionary Learning finds $D=[ v_0 \ldots v_n ]$
and $\theta = [\theta_i \ldots \theta_m]$, such that $x_i = D \theta_i$ and $\Arrowvert \theta_i \Arrowvert_{0} \leq s$. 
\end{definition}

Where the  $\Arrowvert . \Arrowvert_0$ is the $L_0$-norm or sparsity.
If all entries of $\theta_i$ are restricted to be non-negative, we
obtain Sparse-Non-negative Matrix Factorization (SNMF). An optimization version of Dictionary Learning can be written as: 
 \begin{displaymath}
\min_{D \in R^{d,n}}{ \max_{x_i} {\min \Arrowvert \theta_i \Arrowvert_0  : x_i = D\theta_i}} 
\end{displaymath}

In practice, the Dictionary Learning problem is often relaxed to the Lagrangian:
 \begin{displaymath}
 \min \Arrowvert X - D\theta \Arrowvert_2  + \lambda \Arrowvert \theta \Arrowvert_1
\end{displaymath}

Where $X = [x_0 . . . x_m ]$ and $\theta = [\theta_0 . . . \theta_m]$. Several dictionary learning algorithms work by iterating the following two steps.

\begin{enumerate}
 \item Solve the vector selection problem for all vectors $X$. This can be done using your favourite vector selection algorithm, such as basis pursuit. 
 \item Given $X$, the optimization problems is now convex in $D$. Use your favorite method to find $D$.
\end{enumerate}

Using a maximum likelihood formalism, the Method of Optimal Dictionary (MOD) uses the psuedoinverse to compute $D$:

\begin{displaymath}
D^{(i+1)}=X\theta^{(i)^T} (\theta^n \theta^{i^T})^{-1}.
\end{displaymath}

The MOD can be extended to Maximum A-Posteriori probability setting with different priors to take into account preferences in the recovered 
dictionary. Similarly, k-SVD uses an two step iterative process, with a Truncated Singular Value Decomposition to update $D$.
This is done by taking every atom in $D$ and applying SVD to $X$ and $\theta$ restricted to only the columns that that have contribution from that atom.

When $D$ is restricted to be of the form $D = [ B_1, B_2 \ldots B_L ]$ where $B_i$'s are orthonormal matrices, then a more efficient pursuit algorithm
is obtained for the sparse coding stage using a block coordinate relaxation.

\subsection{Dimension Reduction}

The DL step learns a representation $\theta_i$ of the input that lives in a high dimension, 
but we can obtain a lower dimensional representation, since
$\theta_i$ is now readily seen as sparse in the standard orthonormal basis of dimension $n$. 
We can obtain a dimension reduction by using applying a linear operator satisfying the Restricted Isometry Property 
from Compressed Sensing theory.

\begin{definition} 
A linear operator $A$ has the Restricted Isometry Property (RIP) for $s$, iff $\exists \delta_s$ such that:
\begin{equation}
 (1-\delta_{s}) \leq \frac{ \Arrowvert Ax \Arrowvert_2^2}{ \Arrowvert x \Arrowvert_2^2} \leq (1+\delta_{s})
\end{equation}
\end{definition}

\begin{figure}[htbp]
  \centering
    \includegraphics[width=3in, scale=0.5]{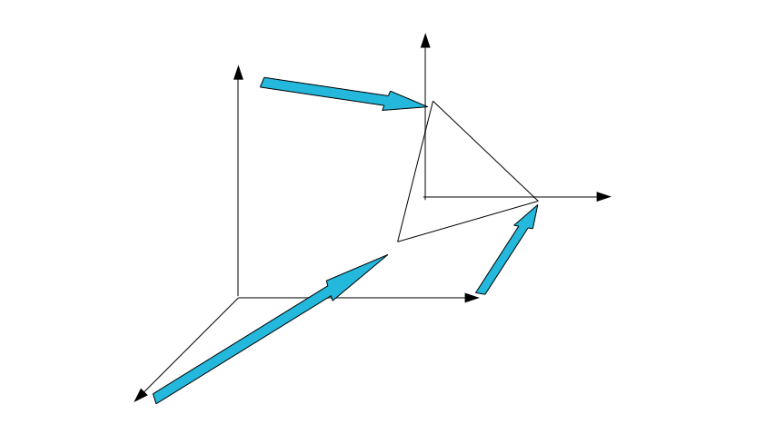}
    \caption{Optimal dimension reduction for a $s=1$ sparse vectors in $d=3$ to vectors in $d=2$.}
\end{figure}

An RIP matrix can compress sparse data while maintaining their approximate relative
distances. This can be seen by considering two $s$-sparse vectors $x_1$ and $x_2$, then:

\begin{equation}
 (1-\delta_{2s}) \leq \frac{ \Arrowvert Ax_1 - Ax_2 \Arrowvert_2^2}{ \Arrowvert x_1 - x_2 \Arrowvert_2^2} \leq (1+\delta_{2s})
\end{equation}

Given an s-sparse vector of dimension $n$, RIP reduces the dimension to $O(s\log(n))$.
Since we are using an RIP matrix, efficient decompression is guaranteed using $L_1$ approximation. 
The data can be recovered exactly using L1 minimization algorithms such as
Basis Pursuit. 

RIP matrices can be obtained probabilistically from matrices with random Gaussian entries. 
Alternatively RIP matrices can be obtained using sparse random matrices \cite{Gilbert2010}. In this paper we follow the latter 
approach. The question of deterministically constructing RIP matrices with similar bounds is still
open.

\subsection{Illustrating the model with simple examples}

\begin{figure}[htb]
\begin{center}
\includegraphics[scale=0.3]{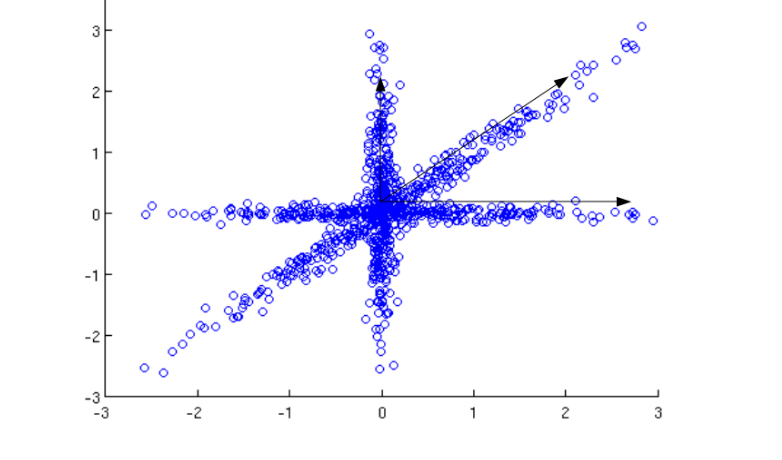}
\end{center}
\caption{Sensory stream data distributed as $s=1$ sparse combinations of the $3$ shown vectors}
\label{fig:awesome_image_2}
\end{figure}

\begin{figure}[htb]
\begin{center}
\includegraphics[scale=0.40]{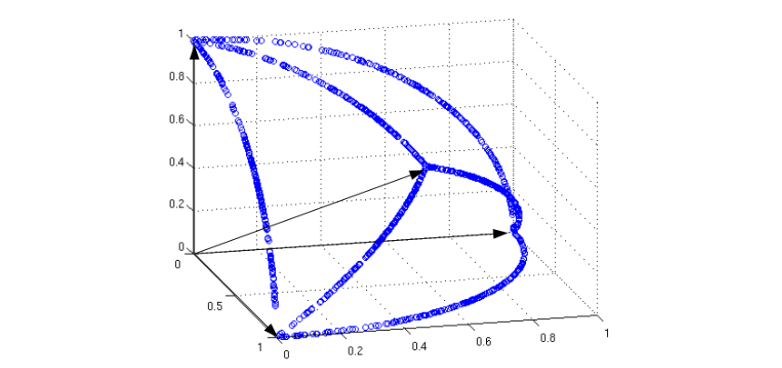}
\end{center}
\caption{Normalized sensory stream data distributed as $s=2$ non-negative sparse combinations of the $4$ shown vectors}
\label{fig:awesome_image_3}
\end{figure}

Let us consider in more detail the working of a single DRDL node. Figure $2$ illustrates a particular distribution that is $s=1$-sparse in the $3$ drawn vectors. 
The Dictionary $D_1$ is

\[ \left( \begin{array}{ccc}
1 & 0 & 1 \\
0 & 1 & 1  \end{array} \right)\]	

The columns of the dictionary learned correspond to the drawn vectors, and the data is expressed simply as a vector with one 
coefficient corresponding to inner product with the closest vector and zero otherwise. This produces an $s=1$ sparse vector in dimension $3$. We can then apply a dimension reduction that preserves distances between those representations. 
The representations correspond to the standard basis in $d=3$.
The best dimension reduction to $d=2$ is then simply the projection of
 the representations onto the plane perpendicular to $(1,1,1)$. Whereby points on the unit basis project to the vertices of a triangle as illustrated
in Figure $1$. This forms the output of the current node.

A slightly more complicated example is $s=2$ in the Dictionary $D_2$ shown bellow. 
Figure $3$ illustrates this distribution for non-negative coefficients. The data was normalized for convenience.

\[ \left( \begin{array}{cccc}
1 & 0 & 0 & 1\\
0 & 1 & 0 & 1\\
0 & 0 & 1 & 1 \end{array} \right)\]

\subsection{Relation between DR and DL}	

There is a symmetry in DRDL due to the fact that DR and DL steps are intimately related.
To show their relationship clearly, we rewrite the two problems with the same variable names.
These variables are only relevant for this section. The two problems can be stated as:

\begin{enumerate}
 \item DL asks for $D$ and $\{x_1 \ldots x_m \}$, given $\{y_1 \ldots y_m \}$, for $Dx_i=y_i$, such that the sparsity $\Arrowvert x_i \Arrowvert_0$ is minimized for a fixed dimension of $y_i$.
 \item DR asks for $D$ and $\{y_1 \ldots y_m \}$, given $\{x_1 \ldots x_m \}$, for $Dx_i=y_i$, such that the dimension of $y_i$'s is minimized for a fixed sparsity $\Arrowvert x_i \Arrowvert_0$.  
\end{enumerate}

In practice, both problems use $L_1$ approximation as a proxy for $L_0$ optimization.
This leads to the following observation

\newtheorem*{observation}{Observation}
\begin{observation} 
The inverse of a DRDL is a DRDL. 
\end{observation}

This means that the space of mappings/functions of our model is the same as it's inverse.
This property will be useful for incorporating feedback.

\subsection{Discussion of Trade-offs in DRDL}	

DRDL can be thought of a memory system ('memory pocket') or a dimension reduction technique for data that can be expressed sparsely in a dictionary.
One parameter trade-off is between $n$, the number of columns in $D$, and $s$, the sparsity of the representation.

On one hand, we note that the DR step puts the data in $O(s log(n))$ dimension. Therefore, if we desire to maximize the reduction in dimension,
increasing $n$ by raising it to a constant power $k$ is comparable to multiplying $s$ by $k$. This means that we much rather
increase the number of columns in the dictionary than the sparsity. On other hand, increasing the number of columns in $D$ forces the columns to be highly correlated. 
Which will become problematic for Basis Pursuit vector selection.

This trade-off highlights the importance of investigating approaches to dictionary learning and vector selection that can go beyond current
results into highly coherent dictionaries. We will further discuss this topic in future papers.

\section{A Hierarchical Sparse Representation}

If we assume a hierarchical architecture modeling the topographic organization of the visual cortex, a singular DRDL element can be 
factorized and expressed as a tree of simpler DRDL elements.
With this architecture we can learn a Hierarchical Sparse Representation
by iterating DRDL elements. 

\subsection{Assumptions of Generative Model}
Our models assumes that data is generated by a hierarchy of spatiotemporal
invariants. At any given level $i$, each node in the generative model is assumed to be
composed of a small number of features $s_i$ . Generation proceeds by recursively
decompressing the pattern from parent nodes then producing patterns to child
nodes. This input is fed to the learning algorithm bellow. 
In this paper we assume that both the topology of generative model and the spatial and 
temporal extent of each node is known. Discussion of algorithms for learning the topology and internal dimensions is
left for future work. 

Consider a simple data stream consisting of a spatiotemporal sequences from a generative model defined above. Figure $1$ shows a potential
learning hierarchy. For simple vision problems, we can consider all dictionaries within a layer as the same. In this paper, processing proceeds
bottom-up the hierarchy only. 

\begin{figure}[htb]
\begin{center}
\includegraphics[scale=0.24]{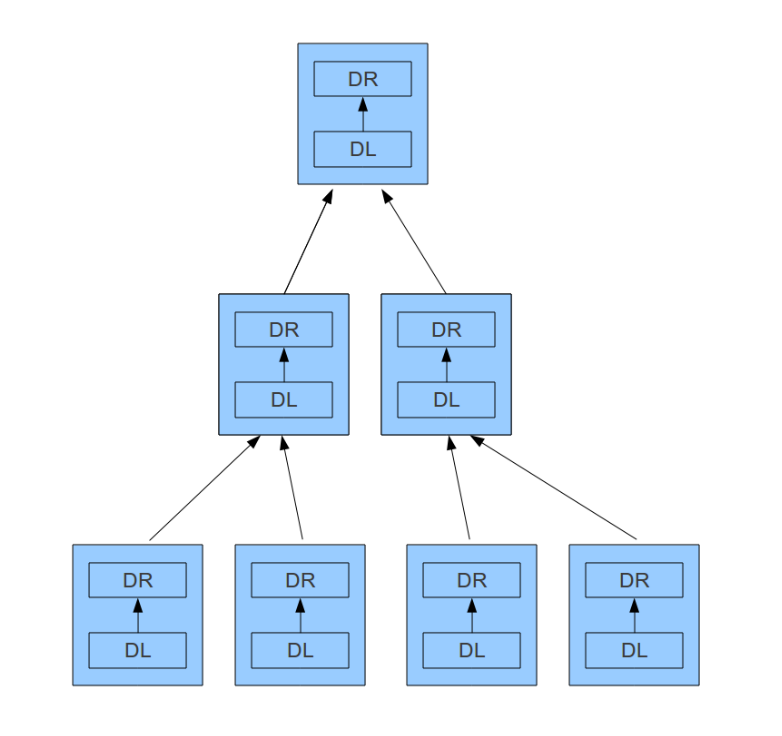}
\end{center}
\caption{A simple 3 layer Hierarchy with no cycles}
\label{fig:awesome_image_4}
\end{figure}

\subsection{Learning Algorithm}
Recursively divide the spatiotemporal signal $x_i$ to obtain a tree representing the known
topographic hierarchy of spatiotemporal blocks. Let $x^0_{i.j}$ be the $j$th block at
level $0$. Then, starting a the bottom of the tree, do:
\begin{enumerate}
 \item Learn a Dictionary $D_{j,k}$ in which the spatiotemporal data $x^k_{i,j}$ can be represented sparsely. This produces a vector of weights $\theta_{i,j,k}$.
 \item Apply dimensionality reduction to the sparse representation to obtain $u_{i,j,k} = A \theta_{i,j,k}$.
 \item Generate $x^{k+1}_{i,j}$ by concatenating vectors $u_{i,l,k}$ for all $l$ that is child of $j$ in at level $k$ in the tree. Replace $k = k + 1$. And now $j$ ranges over elements of level $k$. If k is still less than the depth of the tree, go to Step 1.
\end{enumerate}

Note that in domains such as computer vision, it is reasonable to assume that
all Dictionaries at level $i$ are the same $D_{j,k} = D_k$ . This algorithm attempts
to mirror the generative model. It outputs an inference algorithm that induces
a hierarchy of sparse representations for a given data point. This can be
used to abstract invariant features in the new data. One can then use a supervised 
learning algorithm on top of the invariant features to solve classification
problems.

\subsection{Representation Inference}

For new data points, the representation is obtained, in analogy to the learning algorithm,
recursively dividing the spatiotemporal signal to obtain a tree representing the known
topographic hierarchy of spatiotemporal blocks. The representation is inferred naturally by 
iteratively applying Vector Selection and Compressed Sensing.
For Vector Selection, we employ a common variational technique called Basis Pursuit De-Noising (BPDN), which 
minimizes $\Arrowvert D\theta_i - x_i \Arrowvert_{2}^2 + \lambda \Arrowvert \theta_i \Arrowvert_{1} $. This technique produces optimal results when the sparsity 

\begin{equation}
\Arrowvert \theta \Arrowvert_{0} < \frac{1}{2} + \frac{1}{2C}.
\end{equation}

Where $C$ is the coherence of the dictionary $D$. 

\begin{equation}
C=\max\limits_{k,l} (D^T  D)_{k,l}
\end{equation}

This is a limitation in practice since it is desirable to
have highly coherent dictionaries. Representation inference proceeded by iteratively applying Basis Pursuit and compressing the resulting
sparse representation using the RIP matrix.

\subsection{Mapping between HSR and current models}

This simple, yet powerful, toy model is used as a basis investigation tool for our future work. 
 We abstracted this model out of a need for conceptual simplicity and generality. 
Several models inspired by the neocortex have been proposed, many sharing similar characteristics.
We present a few here and compare to our model. 

H-Max is mapped to HSR by replacing Basis Pursuit for Sparse Approximation with Template Matching. H-Max uses random templates
from the data, whereas HSR uses Dictionary Learning. The 'max' operation in H-Max can be understood in terms of the operation of sparse
coding, which produces local competition among features representing slight variations in the data. Alternatively, the 'max' operation 
can be viewed as a limited form of dimension reduction.

HTM can be mapped to HSR by considering time as another dimension of space.
This way, bounded variable order markov chains can be written as Sparse Approximation of spatiotemporal
vectors representing sequences of spacial vectors in time. HTM constructs a set of spatial coincidences that is automatically shared across time. HSR may do the
same when fed a moving window of spatial sequences. Alternatively, HSR will simulate an HTM by alternating 
between time-only and space-only spatiotemporally extended blocks. In this view a single HTM node is mapped to two layer HSR nodes, one representing time, and the other representing space. 
Unlike HTM, treating time and space on equal footing has the added advantage that the same algorithms used be used for both. HTM with “winner-takes-all” policy can then be mapped to our model by assuming sparsity $s = 1$. HTM
with “distribution on belief states” can be mapped to our model with Template Matching instead of Sparse Approximation in the inference step. Finally, HTM does not leverage RIP dimension reduction step. HTM uses feedback connections for prediction, which is restricted to predictions “forward” in time.
Extending HSR with feedback connections, which accounts for dependency between nodes that are not connected directly,
enables feedback to affect all space-time. 

\subsection{What does HSR offer beyond current models?}

One advantage of our approach is in providing invertibility without a dimensionality blow-up for hierarchically sparse data.
Models falling under Bouvrie et al's framework loose information as you proceed up the hierarchy due to the Pooling operation.
This becomes problematic when extending the models to incorporate feedback. Moreover, this information loss forces an
algorithm designer to hardcode what invariances a particular model must select for (such as translational invariance in H-max). 
On the other hand, invertible models such as HTM suffer from dimensionality blow up, when the number of features learned at a given level
is greater than the input dimension to that level - as is usually the case. 
Dimensionality reduction achieves both a savings in computational resources as well as better noise resilience by avoiding overfitting. 

Dictionary learning represents the data by sparse combinations of dictionary columns. This can be viewed as a $L_0$ or $L_1$ regularization, which provides
noise tolerance and better generalization. This type of regularization is intuitive and is well motivated by neuroscience
 \cite{Olshausen1997, Olshausen1996} and the organization of complex systems. Our approach departs from current models that use
simple template matching and leverages the expressive power of Sparse Approximation. This provides a disciplined prescription for 
learning features at every level. Finally, HSR is a conceptually simple model. This elegance lends itself well to analysis. 

\subsection{Discussion of Trade-offs in HSR}

There are several design decisions when constructing an HSR.
Informally, the hierarchy is useful for reducing the sample complexity and dimensionality of the problem.
For instance consider the simplified case of binary $\{0,1\}$ coefficients and translation invariance (such as vision). An HSR generative model of two layers
will produce $m_2 \choose s_2$ patterns.
Learning this with a single layer HSR would involve learning a 
dictionary of $m_2$ columns and $s_2$ sparsity using $|X|$ samples in dimension $d$. Using two layers, we have the first layer 
 learning a dictionary of size $m_1$ and sparsity $s_1$ using $k*|X|$ samples in dimension $\frac{d}{k}$, the second layer
learns a dictionary of size $m_2$ columns and $s_2$ with $|X|$ samples in dimension $k*O(s_1*\log{m_1}) < d$. 
In effect, by adding a layer, we have divided the problem into two simpler problems in terms of dimensionality and sample complexity.
A more formal and complete discussion will be presented in future work.

\section{Experiments}
In this section we elaborate on preliminary numerical experiments performed with DRDL and HSR on basic standard Machine Learning datasets.
We applied our model to the MNIST and COIL datasets and subsequently used the representation as a feature extraction step for a classification algorithm such as Support Vector Machines (SVM)
or k-Nearest Neighbors (kNN). In practice, additional prior assumptions can be included in our model as discussed in the appendix.

\begin{figure}[htb]
\begin{center}
\includegraphics[scale=0.35]{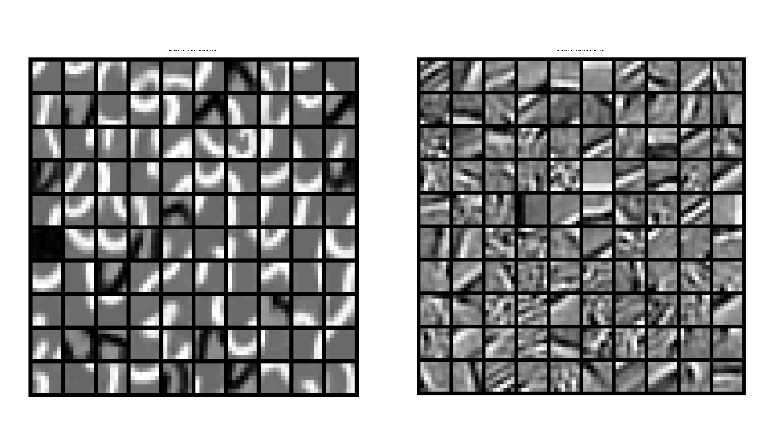}
\end{center}
\caption{Dictionary of the 1st layer of MNIST (left); natural images (right).}
\label{fig:awesome_image_5}
\end{figure}

\subsection{MNIST Results}

\begin{figure}[htb]
\begin{center}
\begin{tabular}{ll}
\textbf{\underline{Method}} & \textbf{\underline{Error}} \% \\
Reconstructive Dictionary Learning & 4.33 \\
Supervised Dictionary Learning & 1.05 \\
k-NN , $l_2$ & 5.0 \\
SVM-Gauss & 1.4 \\
One layer of DRDRL & 1.24 \\
Two layers of HSR & 2.01 \\
\end{tabular}
\end{center}
\caption{A straightforward application of our approach is competitive on the MNIST dataset}
\end{figure}

We applied our model to all pairs of the MNIST data set. 
For the RIP step, we tried random matrices and sparse-random matrices, and tested the efficacy of the approach by reconstructing 
the training data with basis pursuit. 
We used only two layers. After the features are learned we applied a k-NN with $k=3$. 
We refrained from tweaking the initial parameters since we expect our model to work off-shelf. 
For one layer of DRDL we obtained an error rate of $1.24\%$ with a standard deviation of $0.011$.
Using two layers we obtained an error of $2.01\%$ and standard deviation of $0.016$. 

\subsection{COIL Results}

\begin{figure}[htb]
\begin{center}
\begin{tabular}{ll}
\textbf{\underline{Method}} & \textbf{\underline{Classification}} \% \\
One layer of DRDRL & 87.8 \\
SVM & 84.9 \\
Nearest Neighbor & 81.8\\
VTU & 89.9 \\
CNN & 84.8 \\
\end{tabular}
\end{center}
\caption{A straightforward application of our approach is competitive on the COIL dataset}
\end{figure}

We applied our model to all pairs of the COIL-30 (which is a subset of 30 objects out of the entire 100 objects in the COIL datasets).
The data set consists of 72 images for every class. We used only 4 labeled images per class for training. These are taken at equally spaced
angles ($0, 90, 180, 270$). We used the same procedure as before for obtaining and checking the RIP matrix.
We applied a single layer DRDRL, then trained a k-NN with $k=3$.  We also refrained from tweaking the initial parameters. 
We obtained a mean error of $12.2\%$.

\section{Discussion}	

We introduced a novel formulation of an elemental building block 
that could serve as a the bottom-up piece in the common cortical algorithm. This model leads to several interesting theoretical questions.
In the appendix, we illustrate how additional prior assumptions on the generative model can be expressed within our integrated framework. 
Furthermore, this framework can also be extended to address
feedback, attention, action, complimentary learning and the role of time.
In future work, we will closely probe these issues, propose alternative dictionary learning algorithms,
dimension reduction techniques, and incorporate feedback. 

\section*{Acknowledgements}	
We benefited from discussions with several neuroscientists, specifically we would like to thank Dr Linda Hermer.

\bibliographystyle{plain}
\bibliography{References}

\section*{Appendix: Expressing Prior Assumptions}
\begin{figure}
\begin{center}
\includegraphics[scale=0.35]{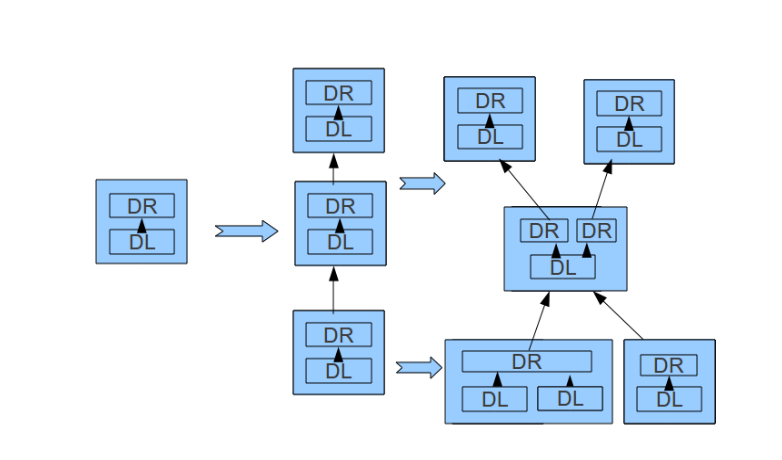}
\end{center}
\caption{Progressive factorizations of a DRDL into a Directed Acyclic Graph (DAG)}
\label{fig:awesome_image_6}
\end{figure}

Additional assumptions reflecting our knowledge of the prior can be explicitly considered within the DRDL and HSR framework.
Additional prior information can be separately encoded at different levels in the hierarchical 
topology and in the Dictionary Learning and Dimension Reduction steps. One type of knowledge is through invariances specified on the data. 
Invariances can be thought of as transformations that map our data to itself. Specifically a set of points $X$ is invariant
under a transformation $I$ iff $I(X)=X$. These assumptions can be encoded to:

\begin{description}
\item \textbf{The Structure of the Model}. The assumption that our generative model can be factored
into a hierarchy is a structural prior. Further factorizations can reflect different structural organizations, such as the topology of the cortex.
Other kinds of assumptions may be imposed as well. For example, in vision, a 
convinient assumption would be that nodes within the same level of hierarchy share the same dictionary. We follow this assumption
in our numerical experiments. This is clearly invalid when dealing with multi-modal sensory data in lower levels. Figure $8$ shows an example of progressive 
factorizations of the model according to prior assumptions.

\item \textbf{The DL Step}. Adding invariance to the Dictionary Learning steps improves the sampling complexity.
For instance, time and space share the property of being shift-invariant. 
One can model the same spatiotemporal block with a single dictionary or with a three level hierarchy of shift-invariant DRDL
reflecting two dimensions of space and one of time. Shift-invariant dictionaries have been studied in the context of learning audio sequences yielding improved performance empirically \cite{Mailhe}. 

\item \textbf {The DR Step}. Imposing invariance selectivity in the dimension reduction step lowers the embedding dimension at the expense of invertibility, similar to H-max.
A more general approach would be to take a middle way of partial selectivity for invariances,
whereby one learns dimension reductions that are lossy for invariances but maintain some variance for a measure of interest.
For example, in vision one can attempt to impose partial selectivity for rotational invariance and scale invariances
by a dimension reduction that pools over small rotations or small scale shifts.
partial-Invariance selectivity can be modeled as the addition of randomness to our generative model. We will explore this direction future work.
\end{description}

\end{document}